\pdfoutput=1
\documentclass[11pt]{article}
\usepackage{acl}
\usepackage{times}
\usepackage{latexsym}
\usepackage[T1]{fontenc}
\usepackage[utf8]{inputenc}
\usepackage{microtype}
\usepackage{inconsolata}
\usepackage{multirow}
\usepackage{graphicx}
\usepackage{algorithm}
\usepackage{algpseudocode}
\usepackage{amsmath} 
\usepackage{amssymb}
\usepackage{colortbl}
\usepackage{url}
\title{Addressing Topic Granularity and Hallucination in Large Language Models for Topic Modelling}

\author{Yida Mu,  Peizhen Bai,  Kalina Bontcheva, Xingyi Song \\
  Department of Computer Science, The University of Sheffield \\
 \texttt{\{y.mu, p.bai, k.bontcheva, x.song\}@sheffield.ac.uk} \\
  }

\begin{document}
\maketitle
\begin{abstract}
Large language models (LLMs) with their strong zero-shot topic extraction capabilities offer an alternative to probabilistic topic modelling and closed-set topic classification approaches. As zero-shot topic extractors, LLMs are expected to understand human instructions to generate relevant and non-hallucinated topics based on the given documents. However, LLM-based topic modelling approaches often face difficulties in generating topics with adherence to granularity as specified in human instructions, often resulting in many near-duplicate topics. Furthermore, methods for addressing hallucinated topics generated by LLMs have not yet been investigated.
In this paper, we focus on addressing the issues of topic granularity and hallucinations for better LLM-based topic modelling.
To this end, we introduce a novel approach that leverages Direct Preference Optimisation (DPO) to fine-tune open-source LLMs, such as Mistral-7B. Our approach does not rely on traditional human annotation to rank preferred answers but employs a reconstruction pipeline to modify raw topics generated by LLMs, thus enabling a fast and efficient training and inference framework. Comparative experiments show that our fine-tuning approach not only significantly improves the LLM's capability to produce more coherent, relevant, and precise topics, but also reduces the number of hallucinated topics. 
\end{abstract}


\section{Introduction}
Topic modelling is a widely-used unsupervised approach that automatically summarises the underlying topics within a large, unlabelled collection of documents \citep{blei2003latent,grootendorst2022bertopic}. It enables the discovery of hidden topics in text data, facilitating easier summarisation and understanding of large datasets \citep{wallach2006topic,vayansky2020review,churchill2022evolution}. 



Large pre-trained language models (LLMs) fine-tuned with feedback on human preferences \citep{touvron2023llama2,OpenAI} provide an entirely new pipeline of topic modelling. For example, LLMs are able to directly produce human-interpretable topics \citep{pham2023topicgpt, mu2024large}, and offer interactive options in topic modelling through user instructions \citep{ouyang2022training,touvron2023llama2}.

\begin{figure}[!t]
    \centering
    \includegraphics[width=1
    \columnwidth]{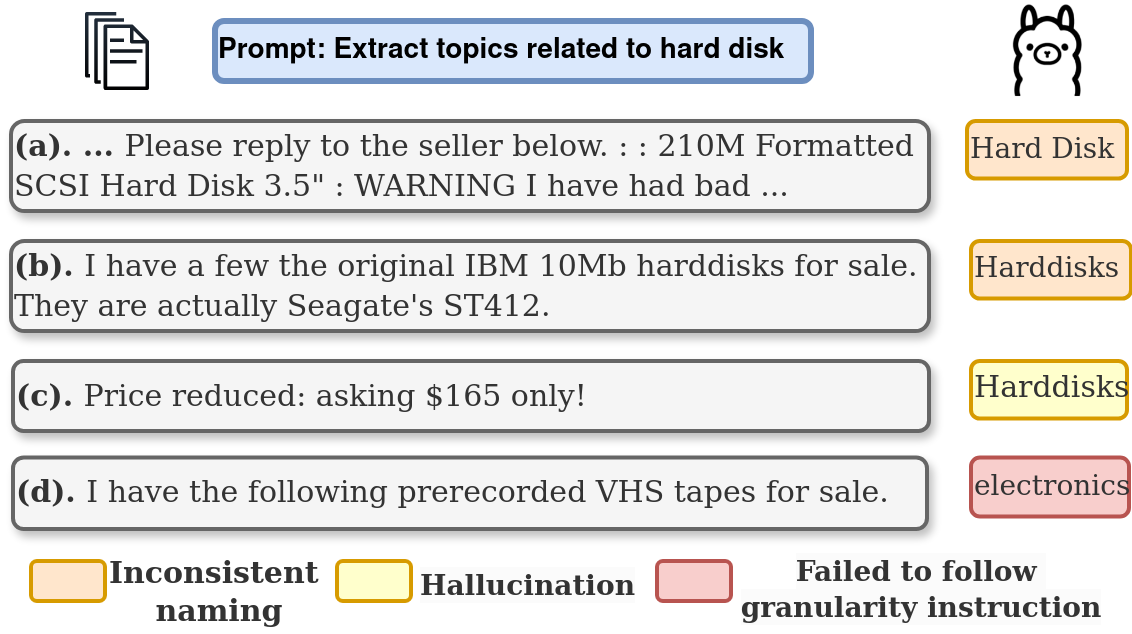} 
    \caption{Four real-world examples consist of the given document (grey), user prompt (blue), and issues associated with LLM-generated topics (see legends with different colours). Examples (a) and (b) demonstrate issues with inconsistent naming; i.e., LLMs tend to generate topics with different formats. Moreover, when prompting LLMs to generate topics related to `hard disk' topics given an unrelated document (examples (c), we observe that LLMs might generate either hallucinated (i.e., `Harddisks') or unwanted topics (i.e., `electronics'). Note that we prompt LLMs to return `No related topics' if there are no related topics in the given text.}
    \label{fig:4_examples} 
\end{figure}

However, employing LLMs in topic modelling introduces new challenges in real-world scenarios. Our experiments reveal that (i) Due to input length limitations, using LLMs for topic extraction is often limited to a single document, and the outputs for different documents display a degree of randomness, generating topics that are near-duplicate (please refer examples (a) \& (b) displayed in Figure~\ref{fig:4_examples}), yet convey highly similar meaning; (ii) LLMs may not always generate topics with the desired granularity, even when granularity instructions are provided \cite{pham2023topicgpt,mu2024large} in the prompt (see examples (d) in Figure \ref{fig:4_examples}). This suggests that current LLMs cannot always generate topics aligned with human instructions, which leads to additional challenges when counting topic statistics and conducting analysis. In addition, our experiments (see \textbf{\S} \ref{hallu_results} and Table \ref{tab:hallucinations}) also reveal the risk of hallucination when provided with more complex prompts such as those incorporating a topic granularity description and seed topics (see samples (c) in Figure \ref{fig:4_examples}). 



The issues of topic granularity, topic naming adherence and risks of hallucinations are vital to the LLM based topic modelling, as they impact the quality of topics prior to any downstream applications (such as temporal topic distribution analysis). In this work, we fine-tune an LLM to directly address these issues making it more suitable for topic modelling tasks. The main contributions are:

\begin{itemize}

    \item To the best of our knowledge, this is the first paper to explore the risks of hallucination across various API-based and open source LLMs within the context of LLM-driven topic modelling. It empirically demonstrates the experimental conditions that can cause LLMs to output hallucinated topics (\textbf{\S} \ref{hallu_results}).
    
    \item We propose an effective reconstruction pipeline for generating Direct Preference Optimisation (DPO) \citep{rafailov2023direct} training samples that do not require human annotations (\textbf{\S} \ref{reconstra_pipeline}).\footnote{Our code and data: \url{https://github.com/GateNLP/TopicLLM_Granularity_Hallucination}}
    
    
    \item Experimental results demonstrate that our fine-tuned LLM (\textbf{TopicMistral}) significantly reduces unlreated topics and hallucinated topics observed in off-the-shelf LLM outputs. Meanwhile, our experiment shows TopicMistral is better at generating topics that are more closely aligned with human-annotated labels (\textbf{\S} \ref{dpo_discussion}). 
    
    \item Finally, we present plug-and-play evaluation protocols designed to measure the quality of topics extracted by LLMs (\textbf{\S} \ref{evaluation_protocols}). 
\end{itemize}

\section{Related Work}
\subsection{LLM-Assisted Topic Modelling}
LLM-Assisted topic modelling refers to the out-of-box application of LLMs, such ChatGPT \citep{OpenAI} or LLaMA \citep{touvron2023llama2}, to assist or enhance the process of traditional probabilistic topic modelling approaches. A typical output format produced by probabilistic topic modelling algorithms is a list of words to represent the topic \citep{blei2003latent,grootendorst2022bertopic}. However, it is very challenging to assign descriptive and meaningful names to the topic identified based on the list \citep{mei2007automatic,mao2012automatic}.
The process of naming a topic, based on its top words, can be undertaken through either manual examination or automated techniques, facilitating a deeper understanding of the underlying thematic content \citep{magatti2009automatic,aletras2014labelling,wan2016automatic}.

Recent work has shown the application of improving BERTopic and LDA, which leverages the text summarising capabilities of LLMs to automatically assign descriptive names to each cluster of words \citep{grootendorst2022bertopic,rijcken2023towards,reuter2024gptopic}. As shown in \citet{rijcken2023towards}, domain experts named topics derived from LDA and found that half of the human-readable names produced by ChatGPT \citep{OpenAI} were deemed useful.

\subsection{LLM-based Topic Modelling}
Previous work on topic modelling with an LLM-based framework is limited. \citet{pham2023topicgpt} introduce a prompt-based topic modelling framework by leveraging the strong capabilities of GPT-4. Similarly, \citet{mu2024large} conduct a battery of experiments on topic modelling with rule-based topic clean approaches. 

To address the topic granularity and naming issues, some rule-based strategies are straightforward to implement but may require domain knowledge; for instance, crafting manual rules to match and merge these topics or employing pre-trained word embeddings (e.g., Sentence Transformer \citep{reimers2019sentence}) for clustering near-duplicate topics based on the semantic similarity \citep{mu2024large}. As an alternative, one might use additional prompt strategies on top of LLM outputs to merge near-duplicate topics under a single topic name \citep{pham2023topicgpt}. However, post-processing could significantly increase the computational complexity and require domain knowledge to craft the post-process rules. 

Moreover, previous studies have neglected the risk of hallucinated topics generated by LLMs; as a result, strategies for mitigating hallucinations in LLM-based topic modelling remain unexplored.

\section{Methodology}

This section describes our methodology for LLM-based topic modelling. \textbf{$\S$}~\ref{sec:zero_shot} introduces the prompting strategies to control topic granularity and adherence to topic naming requirements. \textbf{$\S$}~\ref{sec:fine_tune} describes our fine-tuning strategy to enhance LLM-based topic modelling and hallucination mitigation.

\subsection{Controlling Topic Granularity Though Prompts} \label{sec:zero_shot}

\begin{figure}[!t]
    \centering
    \includegraphics[width=1\columnwidth]{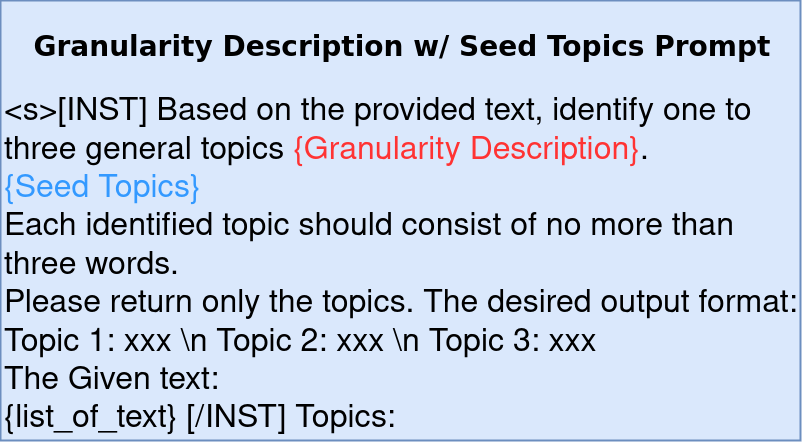} 
    \caption{An example prompt used in our work: Text enclosed by the special tokens `\textit{[/INST]}' denotes the user instruction; \textcolor{red}{red} and \textcolor{blue}{blue} colours denote the Granularity Description and Seed Topics, respectively.}
    \label{fig:promts} 
\end{figure}

We experimented with two prompting strategies (\textbf{Granularity Description} and \textbf{Seed Topics}) to control topic granularity and adherence to topic naming. Our experiments (\textbf{$\S$}~\ref{zero-shot-discussion}) demonstrate that the proposed prompts consistently produce better topics than the baseline prompt, which does not include topic granularity requirements. The description of each prompting strategy is detailed below, and the example prompts are shown in Figure \ref{fig:promts}. Note that we empirically add tokens `Topic:' at the end of the prompt to guide the LLM to generate topics directly, rather than produce extra explanations. 

\begin{itemize}
    \item \textbf{Baseline Prompt (Baseline)}: We start with a baseline prompt without any granularity description and seed topics, which directly instructs LLMs to extract topics from a given text (i.e., the example prompt without the texts in red and blue).

    \item \textbf{Granularity Description (Gran. Desc.)}: An enhanced prompt with  topic granularity requirements, which is aimed at guiding the LLMs to output topics that are at a certain granularity or belong to a given domain (e.g., only output topics related to hard disks). 

    \item \textbf{Seed Topics (Seed Topics)}: Incorporating seed (i.e. example) topics into the prompt provides the LLM with additional guidance on the required topic granularity and naming conventions. For example, we introduce seed topics such as `Hard Disk' to prevent LLMs from generating near-duplicate topics such as `harddisk'. 
\end{itemize}

As detailed below (\textbf{$\S$}~\ref{sec:zero_shot}), we demonstrate that prompt engineering is only partially effective. This motivated our work on uniform naming and topic granularity issues in LLM-based topic modelling. 

In addition, we show that complex prompting strategies also increase the risk of model hallucination. Therefore, we propose a fine-tuning pipeline next. Further details on the hallucination experiments are provided in Section~\ref{hallu_results}.



\subsection{Fine-tuning LLMs Towards to Topic Modelling}
In order to address the limitations of applying off-the-shelf LLMs for topic modelling, we design an improved fine-tuning pipeline for topic modelling through Direct Preference Optimization (DPO) \citep{rafailov2023direct}. In comparison to standard reinforcement learning from human feedback (RLHF \citep{ouyang2022training}), DPO refactors the fine-tuning process with a simple classification loss, thereby eliminating the need for explicit reward.

\paragraph{DPO Fine-tuning Details}
We fine-tune the off-the-shelf LLM (Mistral-7B \citep{jiang2023mistral}) following the DPO pipeline proposed by \citet{rafailov2023direct}. With the introduction of the reference frozen LLM $\pi_{\text{ref}}$, the reference probabilities for accepted topics $y_a$ and rejected topics $y_r$ can be calculated. Then we formulate the DPO maximum likelihood objective for optimizing the LLM fine-tuning $\pi_{\theta}$ as follows:

\begin{equation}
\small
\begin{split}
    \mathcal{L}_\text{DPO}(\pi_{\theta}; \pi_{\text{ref}}) = & -\mathbb{E}_{(x, y_a, y_r)\sim \mathcal{D}}\left[\log \sigma \left(\beta \log \frac{\pi_{\theta}(y_a\mid x)}{\pi_{\text{ref}}(y_a\mid x)} \right. \right. \\
    &\left. \left. - \beta \log \frac{\pi_{\theta}(y_r\mid x)}{\pi_{\text{ref}}(y_r\mid x)}\right)\right],
\end{split}
\end{equation}


\noindent where $\beta$ is a hyperparameter that controls the deviation from $\pi_{\text{ref}}$ to $\pi_{\theta}$ and is set to $0.1$ in our task. $\sigma(\cdot)$ is a sigmoid function. By using the alternative parameterisation mechanism, DPO can fit an implicit reward in the objective, analogous to the reward model in RLHF.

To understand the update mechanism of $\mathcal{L}_\text{DPO}$, the gradient regarding the learnable parameters $\theta$ in $\pi_{\theta}$ is presented as:

\begin{multline*}
    \nabla_\theta \mathcal{L}_\text{DPO}(\pi_\theta;\pi_{\text{ref}}) = \\ -\beta\mathbb{E}_{(x, y_a, y_r) \sim \mathcal{D}} \bigg[\underbrace{\sigma(\hat{r}_\theta(x, y_r) - \hat{r}_\theta (x, y_a))}_\text{higher weight if reward estimate is wrong}
    \\\bigg[\underbrace{\nabla_\theta\log \pi(y_a \mid x)}_\text{increase likelihood of $y_a$} - \underbrace{\nabla_\theta\log\pi(y_r \mid x)}_\text{decrease likelihood of $y_r$}\bigg]\bigg],
\end{multline*}

\noindent where $\hat{r}_\theta(x, y) = \beta \log \frac{\pi_\theta(y \mid x)}{\pi_{\text{ref}}(y \mid x)}$ denotes the implicit reward defined by $\pi_{\theta}$ and $\pi_{\text{ref}}$. This gradient intuitively increases the likelihood of the accepted topics $y_a$ while decreasing the likelihood of rejected topics $y_r$.

\begin{figure}[!t]
    \centering
    \includegraphics[width=1\columnwidth]{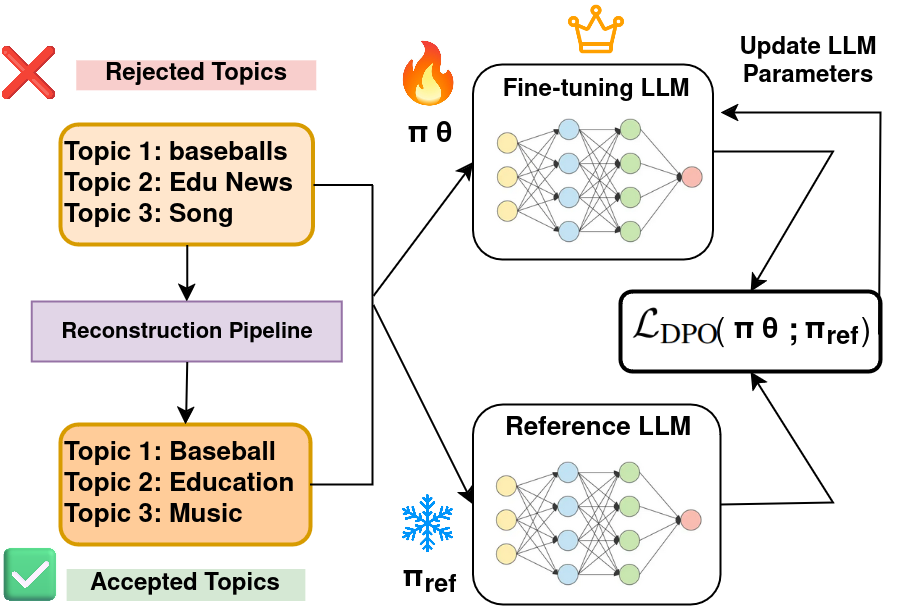} 
    \caption{Our DPO fine-tuning framework. The `reconstruction pipeline' (in purple) denotes the approaches we used to modify the original output from LLMs. We introduce the details of the reconstruction pipeline in Section \ref{reconstra_pipeline}.}
    \label{fig:pipeline} 
\end{figure}

\paragraph{DPO Framework}
The standard pipeline for DPO involves training the LLM to optimise directly for the preference of which `Answer' is most relevant, given two candidate answers. In our case, the format of training samples for DPO fine-tuning comprises three parts, with examples of accepted and rejected topics provided in Figure \ref{fig:pipeline}:
\begin{itemize}
    \item \textbf{Instruction ($x$)}: User prompt including granularity description and seed topics that specify the task requirements.
    \item \textbf{Accepted Topics} ($y_a$): Topics that meet the requirements of the task, including consistent topic names with expected granularity and non-hallucinatory topics.
    \item \textbf{Rejected Topics} ($y_r$): Represents the topics that fail to adhere to the provided instructions.
\end{itemize}

\subsection{Developing Accepted ($y_a$) and Rejected Topics ($y_r$)} \label{sec:fine_tune}
\label{reconstra_pipeline}
%


In classic RLHF scenarios, human annotations are crucial, especially for training the reward model, which serves as the foundation for aligning the model behaviour with human preferences (i.e., `Accepted Topics ($y_a$)') \citep{ouyang2022training}. 

In this study, we introduce a novel data reconstruction pipeline without the need for human annotation for answer ranking purposes. Our approach involves refining the raw topics generated by out-of-shelf LLMs to modify them to meet our instructions. 

Given a set of documents, by summarising the raw topics generated by off-the-shelf LLM, we identify $T$ unique topics, denoted as \( T = \{t_1, t_2, t_3, \ldots, t_n\} \). We consider the top 30 most frequent topics ($T_\text{Sub}$) from $T$ as candidates.

For each of the topics $T_{\text{Sub},i}$ from $T_{\text{Sub}}$, we then use a sentence-transformer\footnote{We use a compact, high-inference-speed base model, namely `all-MiniLM-L6-v2', to map each topic into a 384-dimensional vector space for cosine similarity experiments. This model is available through HuggingFace at: \url{https://huggingface.co/sentence-transformers/all-MiniLM-L6-v2}.} \citep{reimers2019sentence} to identify the topic clusters, which contain near-duplicate topics from $T$, by computing cosine similarity\footnote{Based on exploratory tests, we empirically set the threshold at 0.55 to determine whether a pair of topics are similar, consistent with \citep{pham2023topicgpt}.}. This process results in 30 clusters, which is used to create a matrix for topic matching and replacement. We limit our selection to 30 candidate topics because matching all topics from the extensive list  \(T\) would be overly time-consuming. This limitation aligns with our goal of developing efficient approaches.

For the LLM-generated raw topics of each given document, we proceed to match and replace the raw topics using the matrix we obtained above to generate the `Accepted Topics' (see Figure~\ref{fig:pipeline}). For example, we replace `baseballs' with `Baseball'.

In order to minimise hallucinations in the LLM outputs, we also experiment with creating DPO samples for hallucination mitigation. 
Given documents from the 20NG dataset, covering topics such as Computers and Religion, we prompt the LLM to generate topics related to out-of-distribution domains (e.g., COVID-19). In this case, we expect the LLM will output `\textit{}' rather than hallucinated topics. Therefore, we are able to create accepted topics ($y_a$) by replacing all hallucinated topics with  a`\textit{No Relevant Topics}' answer.

\paragraph{Rejected Topics ($y_r$)}
We use the raw output from the off-the-shelf LLM as rejected topics. Further data cleaning is conducted by filtering out instances where the raw output requires no modification.

Finally, we obtain 2,500 and 900 valid samples for granularity and hallucination fine-tuning, respectively, which are further divided into a DPO training set of 2,800 samples and a DPO validation set of 600 samples.

\section{Experimental Settings}
\subsection{Datasets and Splits}
Following previous work on LLM-based topic modelling \citep{pham2023topicgpt,mu2024large}, we use three publicly available datasets that include annotated topic labels.
\begin{itemize}
    \item \texttt{20 News Groups} (\texttt{20NG}) dataset\footnote{\url{http://qwone.com/~jason/20Newsgroups/}} is an open-domain dataset that contains documents spanning 20 categories, which is widely recognised as a standard benchmark for various NLP downstream tasks.

    \item \texttt{20NG Specific} We also use a subset of the 20NG dataset containing documents specifically related to the `Recreation' news.

    \item \texttt{Wiki} dataset \citep{merity2018regularizing} includes 14k Wikipedia articles that have been annotated into one of 15 general categories.

    \item \texttt{Bills} dataset \citep{adler2018congressional} comprises summaries of 33k bills from the 110th to the 114th U.S. Congresses, annotated with 21 general topics by humans. \footnote{We use the pre-processed \texttt{Wiki} and \texttt{Bills} datasets provided by \citet{pham2023topicgpt}.}
\end{itemize}
We create two training sets for DPO fine-tuning and four different test sets for model evaluation, covering both in-distribution and out-of-distribution settings:

\paragraph{Fine-tuning Sets}
We use 5,000 and 1,000 documents from \texttt{Bills} (10\%) and \texttt{20NG} (5\%) to develop DPO training samples for addressing topic granularity and hallucinations, respectively. 

\paragraph{Test Sets for Topic Granularity}
We use 1,000 documents from each of the \texttt{20NG}, \texttt{Wiki}, and \texttt{Bills} datasets as hold-out sets. Additionally, we extract another 1,000 documents from the \texttt{20NG} dataset, specifically from the `Sports and Recreation' category (namely \texttt{20NG Specific}), to serve as a specific domain dataset for testing LLMs on their ability to generate topics with varying levels of granularity. Note that we do not use documents from \texttt{Wiki} for the DPO fine-tuning.

\paragraph{Tests Sets for Hallucinations}
We use three smaller test sets, each containing 100 documents from the Sports, Technology, and Politics categories of the \texttt{20NG}, respectively. These test sets are strictly excluded from the DPO training sets.
(See Table \ref{tab:hallucinations_2_dpo})

\subsection{Evaluation Protocols}
\label{evaluation_protocols}
Compared to traditional, probabilistic topic models, LLM-based topic modelling directly generates human-interpretable topic names rather than a list of topic words. Therefore, traditional probabilistic topic modelling evaluation metrics such as topic coherence \cite{chang2009reading,newman2010automatic} are not valid for evaluating LLM-based topic modelling. In this section, we propose a set of automatic evaluation metrics to assess the generated topics in the dimensions of naming adherence, human expectation alignment, and risk of hallucination.

\paragraph{Metric 1: Number of Unique Topics (\# of Unique)} is an indicator of the topic naming adherence rate. A better model should produce a smaller number of unique topics because similar yet duplicated topics are merged. However, reducing the number of topics should not diminish the informativeness of the generated topics. Therefore, we also introduce the following metric to assess informativeness.

\paragraph{Metric 2: Similarity Among Top N Topics (Similar N)} is used to evaluate the informativeness of the generated topics. We argue that the generated topics should maximise the semantic differences among each other to convey the most topical information in the documents. The metric measures the average semantic similarity of the top N frequent topics, and is calculated using the following formula ($Topic_x$ is sum of the word embeddings of the topic $x$, we use SentenceBERT \citep{reimers2019sentence} to calculate the word embedding):
\begin{equation}
\footnotesize
\text{Similar $N$} = \frac{\sum_{i=1}^{N-1}\sum_{j=i+1}^{N} \text{Cosine}(Topic_i, Topic_j)}{\frac{N \cdot (N-1)}{2}}
\end{equation}

\paragraph{Metric 3: Mutual Information (MI)} measures the alignment between model-generated topics and human expectations. Specifically, we use MI for quantifying the similarity between topics across two lists: the topics generated by the model and the list of human labelled topics from the datasets.\footnote{Note that we reconstruct the names of categories in 20NG to make them more interpretable. For example, we convert the original label `comp.graphics' into `Computer Graphics'.} The higher Mutual Information score indicates that the topics are better aligned with human expectations. We compute the MI metric using the formula below: (as described in Metric 2, $Topic_x$ is the sum of the word embeddings for topic $x$.)




\noindent Let $L\_1$ and $L\_2$ be two lists of LLM generated topics and human-annotated labels, respectively. For each one-to-one mapping pair ($Topic\_{i}$, $Topic\_{j}$) where $Topic\_{i}$ $\in$ $L\_1$ and $Topic\_{j}$ $\in$ $L\_2$, we compute the cosine\_pairwise($Topic\_{i}$, $Topic\_{j}$).

\begin{equation}
\footnotesize
\begin{split}
     \text{Cosine\_pair}(x) &= \text{Cosine}(Topic_i, Topic_j) \\
    & \noindent \forall \, (Topic_i, Topic_j) \, | \, Topic_i\in L_1 \& Topic_j \in L_2
\end{split}
\end{equation}


\noindent Finally, the MI score over $n$ pairs is calculated as follows::
\begin{equation}
    \text{MI} = \frac{1}{n} \sum_{x=1}^{n} \text{Cosine\_Pair}(x),
\end{equation}

{\noindent where $n$ is the total number of item pairs from $L_1$ and $L_2$, and \text{Cosine\_Pair}(x) is the obtained similarity value for the $x$-th item pair.}

\begin{table*}[!t]
\resizebox{\textwidth}{!}{%
\begin{tabular}{llcccccccccccc}
\hline
\rowcolor[HTML]{FFCE93} 
\cellcolor[HTML]{FFCE93} &
  \cellcolor[HTML]{FFCE93} &
  \multicolumn{3}{c}{\cellcolor[HTML]{FFCE93}\textbf{20NG Specific}} &
  \multicolumn{3}{c}{\cellcolor[HTML]{FFCE93}\textbf{20NG}} &
  \multicolumn{3}{c}{\cellcolor[HTML]{FFCE93}\textbf{Bills}} &
  \multicolumn{3}{c}{\cellcolor[HTML]{FFCE93}\textbf{Wikipedia}} \\ \cline{3-14} 
\rowcolor[HTML]{FFCE93} 
\multirow{-2}{*}{\cellcolor[HTML]{FFCE93}\textbf{Models}} &
  \multirow{-2}{*}{\cellcolor[HTML]{FFCE93}\textbf{Prompts}} &
  \cellcolor[HTML]{FFCE93}\textbf{\#} &
  \textbf{\begin{tabular}[c]{@{}c@{}}Similar\\ N\%\end{tabular}} &
  \cellcolor[HTML]{FFCE93}\textbf{MI} &
  \textbf{\#} &
  \textbf{\begin{tabular}[c]{@{}c@{}}Similar\\ N\%\end{tabular}} &
  \cellcolor[HTML]{FFCE93}\textbf{MI} &
  \textbf{\#} &
  \textbf{\begin{tabular}[c]{@{}c@{}}Similar\\ N\%\end{tabular}} &
  \cellcolor[HTML]{FFCE93}\textbf{MI} &
  \textbf{\#} &
  \textbf{\begin{tabular}[c]{@{}c@{}}Similar\\ N\%\end{tabular}} &
  \textbf{MI} \\ \hline
\rowcolor[HTML]{FFFFFF} 
LDA &
  \multicolumn{1}{c}{\cellcolor[HTML]{FFFFFF}\textbf{-}} &
  \textbf{-} &
  \cellcolor[HTML]{FFFFFF}0.136 &
  \cellcolor[HTML]{FFFFFF}0.298 &
  \textbf{-} &
  0.160 &
  0.202 &
  \textbf{-} &
  \cellcolor[HTML]{FFFFFF}0.156 &
  \cellcolor[HTML]{FFFFFF}0.254 &
  \textbf{-} &
  \cellcolor[HTML]{FFFFFF}0.125 &
  \cellcolor[HTML]{FFFFFF}0.284 \\
\rowcolor[HTML]{FFFFFF} 
BERTopic &
  \multicolumn{1}{c}{\cellcolor[HTML]{FFFFFF}\textbf{-}} &
  \textbf{-} &
  \cellcolor[HTML]{FFFFFF}0.170 &
  \cellcolor[HTML]{FFFFFF}0.217 &
  \textbf{-} &
  0.171 &
  0.286 &
  \textbf{-} &
  \cellcolor[HTML]{FFFFFF}0.171 &
  \cellcolor[HTML]{FFFFFF}0.296 &
  \textbf{-} &
  \cellcolor[HTML]{FFFFFF}0.135 &
  \cellcolor[HTML]{FFFFFF}0.288 \\ \hline
\rowcolor[HTML]{EFEFEF} 
\multicolumn{14}{c}{\cellcolor[HTML]{EFEFEF}\textbf{Zero-shot LLM-based Topic Modelling}} \\ \hline
\rowcolor[HTML]{FFFFFF} 
GPT-3.5 &
  \cellcolor[HTML]{FFFFFF}Baseline &
  \cellcolor[HTML]{FFFFFF}2,625 &
  \cellcolor[HTML]{FFFFFF}0.191 &
  \cellcolor[HTML]{FFFFFF}0.284 &
  \cellcolor[HTML]{FFFFFF}3,058 &
  0.178 &
  0.210 &
  \cellcolor[HTML]{FFFFFF}2,426 &
  0.210 &
  0.292 &
  2,367 &
  0.134 &
  0.313 \\
\rowcolor[HTML]{FFFFFF} 
GPT-3.5 &
  \cellcolor[HTML]{FFFFFF}Gran. Desc. &
  \cellcolor[HTML]{FFFFFF}2,414 &
  \cellcolor[HTML]{FFFFFF}0.197 &
  \cellcolor[HTML]{FFFFFF}0.316 &
  \cellcolor[HTML]{FFFFFF}3,024 &
  0.171 &
  0.221 &
  \cellcolor[HTML]{FFFFFF}- &
  - &
  - &
  - &
  - &
  - \\
\rowcolor[HTML]{FFFFFF} 
GPT-3.5 &
  \cellcolor[HTML]{FFFFFF}Seed Topics &
  \cellcolor[HTML]{FFFFFF}2,515 &
  \cellcolor[HTML]{FFFFFF}0.181 &
  \cellcolor[HTML]{FFFFFF}0.315 &
  \cellcolor[HTML]{FFFFFF}3,026 &
  0.174 &
  0.224 &
  1,838 &
  0.193 &
  0.333 &
  2,017 &
  0.163 &
  0.339 \\
\rowcolor[HTML]{FFFFFF} 
LLaMA-7B &
  Baseline &
  2,185 &
  0.136 &
  0.256 &
  2,703 &
  0.105 &
  0.283 &
  2,358 &
  0.115 &
  0.305 &
  2,360 &
  0.113 &
  0.273 \\
\rowcolor[HTML]{FFFFFF} 
LLaMA-7B &
  Gran. Desc. &
  1,965 &
  0.159 &
  0.275 &
  2,646 &
  0.168 &
  0.290 &
  - &
  - &
  - &
  - &
  - &
  - \\
\rowcolor[HTML]{FFFFFF} 
LLaMA-7B &
  Seed Topics &
  1,440 &
  0.166 &
  0.328 &
  1,930 &
  0.137 &
  0.357 &
  1,816 &
  0.205 &
  0.364 &
  2,244 &
  0.131 &
  0.311 \\
\rowcolor[HTML]{FFFFFF} 
LLaMA-13B &
  \cellcolor[HTML]{FFFFFF}Baseline &
  \cellcolor[HTML]{FFFFFF}2,421 &
  \cellcolor[HTML]{FFFFFF}0.198 &
  \cellcolor[HTML]{FFFFFF}0.302 &
  2,909 &
  0.175 &
  0.266 &
  2,261 &
  0.137 &
  0.300 &
  2,098 &
  0.143 &
  0.331 \\
\rowcolor[HTML]{FFFFFF} 
\cellcolor[HTML]{FFFFFF}LLaMA-13B &
  \cellcolor[HTML]{FFFFFF}Gran. Desc. &
  \cellcolor[HTML]{FFFFFF}1,863 &
  \cellcolor[HTML]{FFFFFF}0.180 &
  \cellcolor[HTML]{FFFFFF}0.395 &
  2,498 &
  0.177 &
  0.316 &
  - &
  - &
  - &
  - &
  - &
  - \\
\rowcolor[HTML]{FFFFFF} 
\cellcolor[HTML]{FFFFFF}LLaMA-13B &
  \cellcolor[HTML]{FFFFFF}Seed Topics &
  \cellcolor[HTML]{FFFFFF}2,003 &
  \cellcolor[HTML]{FFFFFF}0.182 &
  \cellcolor[HTML]{FFFFFF}0.379 &
  2,406 &
  0.176 &
  0.314 &
  2,096 &
  0.189 &
  0.328 &
  1,669 &
  0.107 &
  0.380 \\
\rowcolor[HTML]{FFFFFF} 
Mistral-7B &
  Baseline &
  1,949 &
  0.212 &
  0.321 &
  2,268 &
  0.181 &
  0.282 &
  2,250 &
  0.186 &
  0.291 &
  \cellcolor[HTML]{FFFFFF}2,186 &
  \cellcolor[HTML]{FFFFFF}0.139 &
  \cellcolor[HTML]{FFFFFF}0.312 \\
\rowcolor[HTML]{FFFFFF} 
Mistral-7B &
  Gran. Desc. &
  1,759 &
  0.201 &
  0.363 &
  2,318 &
  0.180 &
  0.298 &
  \cellcolor[HTML]{FFFFFF}- &
  \cellcolor[HTML]{FFFFFF}- &
  \cellcolor[HTML]{FFFFFF}- &
  \cellcolor[HTML]{FFFFFF}- &
  \cellcolor[HTML]{FFFFFF}- &
  \cellcolor[HTML]{FFFFFF}- \\
\rowcolor[HTML]{FFFFFF} 
Mistral-7B &
  Seed Topics &
  2,098 &
  0.200 &
  0.360 &
  2,646 &
  0.190 &
  0.287 &
  \cellcolor[HTML]{FFFFFF}2,177 &
  \cellcolor[HTML]{FFFFFF}0.203 &
  \cellcolor[HTML]{FFFFFF}0.311 &
  \cellcolor[HTML]{FFFFFF}1,939 &
  \cellcolor[HTML]{FFFFFF}0.147 &
  \cellcolor[HTML]{FFFFFF}0.342 \\ \hline
\rowcolor[HTML]{EFEFEF} 
\multicolumn{14}{c}{\cellcolor[HTML]{EFEFEF}\textbf{Zero-shot LLM-based Topic Modelling + Post-processing}} \\ \hline
\rowcolor[HTML]{FFFFFF} 
TopicGPT &
  \multicolumn{1}{c}{\cellcolor[HTML]{FFFFFF}-} &
  - &
  - &
  0.396 &
  - &
  - &
  0.286 &
  - &
  - &
  0.315 &
  - &
  - &
  0.388 \\
\rowcolor[HTML]{FFFFFF} 
\citet{mu2024large} &
  \multicolumn{1}{c}{\cellcolor[HTML]{FFFFFF}-} &
  1,865 &
  0.132 &
  0.355 &
  2,605 &
  0.143 &
  0.272 &
  2,139 &
  0.203 &
  0.308 &
  1,890 &
  0.139 &
  0.338 \\ \hline
\rowcolor[HTML]{EFEFEF} 
\multicolumn{14}{c}{\cellcolor[HTML]{EFEFEF}\textbf{Zero-shot LLM-based Topic Extraction + DPO Fine-tuning}} \\ \hline
\rowcolor[HTML]{FFFFFF} 
\textbf{\begin{tabular}[c]{@{}l@{}}Topic\\ Mistral\end{tabular}} &
  \cellcolor[HTML]{FFFFFF}Seed Topics &
  \textbf{116} &
  \textbf{0.156} &
  \textbf{0.577} &
  \textbf{433} &
  \textbf{0.112} &
  \textbf{0.443} &
  \textbf{192} &
  \textbf{0.104} &
  \textbf{0.485} &
  \textbf{289} &
  \textbf{0.103} &
  \textbf{0.517} \\
\rowcolor[HTML]{FFFFFF} 
\cellcolor[HTML]{FFFFFF}\textbf{\begin{tabular}[c]{@{}l@{}}Topic\\ Mistral\end{tabular}} &
  \cellcolor[HTML]{FFFFFF}\begin{tabular}[c]{@{}l@{}}Seed Topics\\ Dynamic\end{tabular} &
  \textbf{85} &
  \textbf{0.147} &
  \textbf{0.531} &
  \textbf{373} &
  \textbf{0.112} &
  \textbf{0.449} &
  \textbf{192} &
  \textbf{0.103} &
  \textbf{0.488} &
  \textbf{219} &
  \textbf{0.101} &
  \textbf{0.534} \\ \hline
\end{tabular}%
}
\caption{Topic granularity evaluation results across all metrics, LLMs, and prompt strategies. `\textbf{\#}' denotes `\# of Unique'. TopicMistral' denotes the new fine-tuned Mistral-7B model. For MI, higher values \textbf{$\uparrow$} indicate better performance, while lower values \textbf{$\downarrow$} are better for \# of Unique Topics and Similar $N$.
Due to different experimental settings, we are unable to directly compare the number of unique topics and Similar $N$' between TopicGPT and TopicMistral. }
\label{tab:evaluations_all_results}
\end{table*}

\subsection{Implementation Details}
\paragraph{Fine-Tuning}
We use Mistral-7B \citep{jiang2023mistral} as the base model for the DPO fine-tuning experiments. Hence we name our fine-tuned model as \textbf{TopicMistral}.
The choice of Mistral-7B is driven by practical considerations, such as the availability of computational resources. Furthermore, we focus exclusively on open-source models, as fine-tuning and sharing models based on API-dependent LLMs, such as ChatGPT and Claude, is not feasible.

We conduct DPO fine-tuning following the methodology introduced in \citet{rafailov2023direct}.\footnote{Our approach adheres to the standard DPO training pipeline via Hugging Face Transformer Reinforcement Learning (TRL) library: \url{https://huggingface.co/docs/trl/main/en/dpo_trainer}.} Considering the substantial GPU memory requirement of up to 70 GB for fully fine-tuning LLMs with the 7B size in the full precision mode, which is beyond our computational resources; we opt for a Parameter-Efficient Fine-Tuning (PEFT) method, specifically QLoRA \citep{dettmers2023qlora}, available through the Hugging Face PEFT library.\footnote{\url{https://huggingface.co/docs/peft/en/index}.}


We set the deviation controlling hyperparameter $\beta$ to 0.1, learning rate to 4e-5, with a batch size of 2 and 200 warm-up steps. All experiments are executed on a single Nvidia A100 GPU equipped with 40GB of memory. However, given the efficiency of the PEFT approach, our experimental setup is applicable on any commercial GPU with at least 12GB memory.

\begin{algorithm}
\small
\caption{Dynamic Seed Topics}
\begin{algorithmic}[1]
\State \textbf{Input:} \texttt{Docs, N}, 
\State \textbf{Input:} \texttt{Seed\_Topics $\gets$ [initial topics]}, 
\State \textbf{Function:} \texttt{Get\_Top10(topic\_list)}
\State \textbf{Function:} \texttt{TopicMistral(document, Seed\_Topics)}
\State $\texttt{Topic\_List} \gets [ ]$

\For{$i \gets 0$ \textbf{to} $\text{len}(\texttt{Docs})-1$}
    \If{$i > \texttt{N}$}
        \State $\texttt{Seed\_Topics} \gets \texttt{Get\_Top10}(\texttt{Topic\_List})$
    \EndIf
    \State $T\_i \gets \texttt{TopicMistral}(\texttt{Docs}[i], \texttt{Seed\_Topics})$
    \State \texttt{Topic\_List}.append($T\_i$)
\EndFor
\State \Return \texttt{Topic\_List}
\end{algorithmic}
\label{dynamic_algorithm}
\end{algorithm}

\paragraph{Dynamic Seed Topics:} As mentioned in Section~\ref{sec:zero_shot}, one of the primary functions of using seed topics is to control topic granularity. However, manually setting an optimal set of seed topics for an unseen corpus is non-trivial. Hence, we introduce Dynamic Seed Topics, where the list of seed topics is updated dynamically based on the statistics of the topics generated by TopicMistral. 
The pseudo-code for Dynamic Seed Topics is presented in Algorithm~\ref{dynamic_algorithm}. The pipeline first initialises with a set of initial seed topics (\texttt{Seed\_Topics}). After generating topics for N documents (\texttt{Docs}) using an TopicMistral (\texttt{TopicMistral}), the seed topics are replaced with the top 10 most frequent topics (\texttt{Get\_Top10}) from the generated topics (\texttt{Topic\_List}). In our experiment, N is set to 20.

\subsection{Baselines}
This paper evaluates our fine-tuned model against 8 baselines.

\begin{itemize}
    \item \textbf{GPT-3.5 (GPT)}\footnote{\url{https://platform.openai.com/docs/models/gpt-3-5}} is one of the most popular API-based LLM developed by OpenAI, renowned for its strong ability to generate human-like text.

      \item \textbf{LLaMA-2-7B \& LLaMA-2-13B} \citep{touvron2023llama2} represent advanced versions of LLaMA-7B \citep{touvron2023llama}, featuring more training resources for enhanced performance.

    \item \textbf{TopicGPT}: \citet{pham2023topicgpt} uses a post-processing approach by adding additional prompts on the top of the LLM's outputs to merge near-duplicate topics. 
    We adapt their prompts to fit the Mistral-7B model for a fair comparison.

    \item \textbf{\citet{mu2024large}} use manually generated rules to merge near-duplicate topics to solve topic naming issues. The final topics list is obtained by adding an additional prompt on the list of deduplicated topics.

    \item \textbf{Mistral-7B} \citep{jiang2023mistral} For reference, we also evaluate the off-the-shelf Mistral-7B, which uses grouped-query attention for quick inference and incorporates sliding window attention to efficiently process sequences of mixed lengths at a faster inference speed

    \item{ \textbf{LDA} and \textbf{BERTopic}} For reference, we also compare against two widely-used topic modelling approaches: LDA \citep{blei2003latent} and BERTopic \citep{grootendorst2022bertopic}. Since the outputs of LDA and BERTopic are unnamed, we employ LLM\footnote{Specifically, we use the GPT-3.5 Turbo model via the OpenAI API.} to assign names to topics based on a list of representative words for each topic. For example, we prompt the LLM to assign a name for a topic given a list of words, which represents the typical output format of LDA and BERTopic. 
    For direct comparison with LLMs, we also transfer our automatic evaluation protocols to LDA and BERTopic. We directly calculate the `Similar N' metric using the names of the top 10 topics generated by ChatGPT. Similarly, we use names generated by ChatGPT to represent each document in order to assess the `Mutual Information' metric.
\end{itemize}




\section{Results and Analysis}

\begin{table*}[!t]
\resizebox{\textwidth}{!}{%
\begin{tabular}{lclllclllclllclll}
\hline
\rowcolor[HTML]{FFCE93} 
\textbf{Models} &
  \textbf{L-7B} &
  \multicolumn{1}{c}{\cellcolor[HTML]{FFCE93}\textbf{L-13B}} &
  \multicolumn{1}{c}{\cellcolor[HTML]{FFCE93}\textbf{M-7B}} &
  \multicolumn{1}{c}{\cellcolor[HTML]{FFCE93}\textbf{GPT}} &
  \textbf{L-7B} &
  \multicolumn{1}{c}{\cellcolor[HTML]{FFCE93}\textbf{L-13B}} &
  \multicolumn{1}{c}{\cellcolor[HTML]{FFCE93}\textbf{M-7B}} &
  \multicolumn{1}{c}{\cellcolor[HTML]{FFCE93}\textbf{GPT}} &
  \textbf{L-7B} &
  \multicolumn{1}{c}{\cellcolor[HTML]{FFCE93}\textbf{L-13B}} &
  \multicolumn{1}{c}{\cellcolor[HTML]{FFCE93}\textbf{M-7B}} &
  \multicolumn{1}{c}{\cellcolor[HTML]{FFCE93}\textbf{GPT}} &
  \textbf{L-7B} &
  \multicolumn{1}{c}{\cellcolor[HTML]{FFCE93}\textbf{L-13B}} &
  \multicolumn{1}{c}{\cellcolor[HTML]{FFCE93}\textbf{M-7B}} &
  \multicolumn{1}{c}{\cellcolor[HTML]{FFCE93}\textbf{GPT}} \\ \hline
\rowcolor[HTML]{EFEFEF} 
\textbf{Docs.} &
  \multicolumn{4}{c}{\cellcolor[HTML]{EFEFEF}Sports News} &
  \multicolumn{4}{c}{\cellcolor[HTML]{EFEFEF}Sports News} &
  \multicolumn{4}{c}{\cellcolor[HTML]{EFEFEF}Sports News} &
  \multicolumn{4}{c}{\cellcolor[HTML]{EFEFEF}Sports News} \\ \cline{2-17} 
\rowcolor[HTML]{EFEFEF} 
\textbf{Gran. Desc.} &
  \multicolumn{4}{c}{\cellcolor[HTML]{EFEFEF}COVID-19} &
  \multicolumn{4}{c}{\cellcolor[HTML]{EFEFEF}Vehicles} &
  \multicolumn{4}{c}{\cellcolor[HTML]{EFEFEF}COVID-19} &
  \multicolumn{4}{c}{\cellcolor[HTML]{EFEFEF}Vehicles} \\ \cline{2-17} 
\rowcolor[HTML]{EFEFEF} 
\textbf{Seed Topics} &
  \multicolumn{4}{c}{\cellcolor[HTML]{EFEFEF}Side effects \& Effective} &
  \multicolumn{4}{c}{\cellcolor[HTML]{EFEFEF}Autos \& Motorcycles} &
  \multicolumn{4}{c}{\cellcolor[HTML]{EFEFEF}No seed} &
  \multicolumn{4}{c}{\cellcolor[HTML]{EFEFEF}No seed} \\ \hline
\textbf{Adherence\%} &
  \multicolumn{1}{l}{0\%} &
  32\% &
  50\% &
  100\% &
  \multicolumn{1}{l}{0\%} &
  7\% &
  1\% &
  100\% &
  \multicolumn{1}{l}{{\color[HTML]{333333} 0\%}} &
  {\color[HTML]{333333} 18\%} &
  {\color[HTML]{333333} 29\%} &
  {\color[HTML]{333333} 100\%} &
  \multicolumn{1}{l}{{\color[HTML]{333333} 0\%}} &
  3\% &
  0\% &
  99\% \\
\textbf{Hallu.\%} &
  \multicolumn{1}{l}{14\%} &
  29\% &
  5\% &
  0\% &
  \multicolumn{1}{l}{3\%} &
  1\% &
  9\% &
  0\% &
  \multicolumn{1}{l}{{\color[HTML]{333333} 49\%}} &
  {\color[HTML]{333333} 68\%} &
  {\color[HTML]{333333} 13\%} &
  {\color[HTML]{333333} 0\%} &
  \multicolumn{1}{l}{{\color[HTML]{333333} 11\%}} &
  2\% &
  {\color[HTML]{333333} 2\%} &
  1\% \\
\textbf{Aligned\%} &
  \multicolumn{1}{l}{86\%} &
  39\% &
  45\% &
  0\% &
  \multicolumn{1}{l}{97\%} &
  92\% &
  90\% &
  0\% &
  \multicolumn{1}{l}{{\color[HTML]{333333} 51\%}} &
  {\color[HTML]{333333} 14\%} &
  {\color[HTML]{333333} 58\%} &
  {\color[HTML]{333333} 0\%} &
  \multicolumn{1}{l}{{\color[HTML]{333333} 89\%}} &
  95\% &
  {\color[HTML]{333333} 98\%} &
  0\% \\ \hline
\end{tabular}%
}
\caption{ LLM Hallucination Evaluation Results across different combinations of granularity descriptions and seed topics. L-7B and M-7B denote LLaMA-7B and Mistral 7B respectively. For Adherence\%, higher values \textbf{$\uparrow$} indicate better performance, while lower values \textbf{$\downarrow$} are better for Hallucination\% and Aligned\%. The sum of the three metrics should be 100\%.}
\label{tab:hallucinations}
\end{table*}
\paragraph{Off-The-Shelf LLMs}
\label{zero-shot-discussion}
Similar to \citet{pham2023topicgpt} and \citet{mu2024large}, we observe that all off-the-shelf LLMs prompted without seed topics, tend to produce a large number of near-duplicate topics (See Table \ref{tab:evaluations_all_results}). 
When including either granularity descriptions or seed topics, or both, most LLMs demonstrate a slight improvement in their ability to generate fewer near-duplicate topics. However, challenges with inconsistent topic naming persist (e.g., Music, Song, and Music Production). A higher number of inconsistently named topics results in higher \# of Unique and Similar $N$ scores across all off-the-shelf LLMs. This indicates that current LLMs are not well suited to zero-shot topic modelling, underscoring the significance of our research. 

\paragraph{LDA and BERTopic} On the other hand, existing topic modelling approaches such as LDA and BERTopic, with assistance from GPT for topic labelling and summarisation, can achieve similar performance to most off-the-shelf LLMs. This finding is aligned with prior work on LLM-assisted topic modelling \citep{rijcken2023towards,reuter2024gptopic,chang2024enhanced}.

\paragraph{Post-processing Approaches} Our findings on off-the-shelf LLMs highlight the motivation behind prior work, such as that of \citet{pham2023topicgpt} and \citet{mu2024large}, which leverage post-processing methods (such as rule-based topic cleaning and additional prompting to mitigate the generation of inconsistent naming and near-duplicate topics by off-the-shelf LLMs. Note that while post-processing can be also applied to the output of TopicMistral, we believe that this process will be considerably simpler because our LLM produces significantly fewer unique topics (e.g., 90\% less unique topics generated by TopicMistral on \texttt{Bills}).

\paragraph{TopicMistral v.s. Off-The-Shelf Mistral}
\label{dpo_discussion}
A notable observation is that our fine-tuned TopicMistral model reliably yields significantly fewer inconsistent naming and near-duplicate topics across all datasets compared to off-the-shelf LLMs (e.g., $289$ < $1,939$ on \texttt{Bills}).
Note that the lower Similar $N$ values indicate that our model can generate more diverse top $N$ topics (e.g., $0.104$ < $0.203$ on \texttt{Bills}).
Besides, we observe significant improvement in the MI (taking \texttt{Wiki} as an example, the MI value has improved from $0.342$ to $0.534$), which suggests that the fine-tuned TopicMistral model excels in generating topics that align more closely with the original labels.
Considering that the \texttt{Wiki} dataset is not included in the DPO training samples, we argue that our LLM fine-tuning strategies can be generalised to other out-of-distribution datasets.

Meanwhile, our comparative experiments (See Table \ref{tab:hallucinations_2_dpo}) reveal that our fine-tuning approach has significantly reduced the number of hallucinated topics across different adversarial prompt settings. Detailed discussion provided in \textbf{\S} \ref{hallu_results}.

\paragraph{Dynamic List of Seed Topics}
For all datasets, the use of dynamic seed topics can further reduce the number of unique topics with less similar $N$ values, i.e., generating more diverse top $N$ topics. However, this approach may yield slightly worse MI metrics on domain-specific datasets, as seen in \texttt{20N Specific} ($0.147$ < $0.156$). Furthermore, our findings highlight the advantages of using a dynamic list of seed topics when the user is not familiar with the domain of the given set of documents.

\section{Topic Hallucinations}

\begin{table*}[!t]
\small
\resizebox{\textwidth}{!}{%
\begin{tabular}{lllllllllllll}
\hline
\rowcolor[HTML]{FFCE93} 
\textbf{Models} &
  \multicolumn{1}{c}{\cellcolor[HTML]{FFCE93}\textbf{M-7B}} &
  \multicolumn{1}{c}{\cellcolor[HTML]{FFCE93}\textbf{TM-7B}} &
  \multicolumn{1}{c}{\cellcolor[HTML]{FFCE93}\textbf{M-7B}} &
  \multicolumn{1}{c}{\cellcolor[HTML]{FFCE93}\textbf{TM-7B}} &
  \multicolumn{1}{c}{\cellcolor[HTML]{FFCE93}\textbf{M-7B}} &
  \multicolumn{1}{c}{\cellcolor[HTML]{FFCE93}\textbf{TM-7B}} &
  \multicolumn{1}{c}{\cellcolor[HTML]{FFCE93}\textbf{M-7B}} &
  \multicolumn{1}{c}{\cellcolor[HTML]{FFCE93}\textbf{TM-7B}} &
  \multicolumn{1}{c}{\cellcolor[HTML]{FFCE93}\textbf{M-7B}} &
  \multicolumn{1}{c}{\cellcolor[HTML]{FFCE93}\textbf{TM-7B}} &
  \multicolumn{1}{c}{\cellcolor[HTML]{FFCE93}\textbf{M-7B}} &
  \multicolumn{1}{c}{\cellcolor[HTML]{FFCE93}\textbf{TM-7B}} \\ \hline
\rowcolor[HTML]{EFEFEF} 
\textbf{Docs.} &
  \multicolumn{4}{c}{\cellcolor[HTML]{EFEFEF}Sports News} &
  \multicolumn{4}{c}{\cellcolor[HTML]{EFEFEF}Computer News} &
  \multicolumn{4}{c}{\cellcolor[HTML]{EFEFEF}Politics News} \\ \cline{2-13} 
\rowcolor[HTML]{EFEFEF} 
\cellcolor[HTML]{EFEFEF}\textbf{Gran. Desc.} &
  \multicolumn{2}{c}{\cellcolor[HTML]{EFEFEF}COVID-19} &
  \multicolumn{2}{c}{\cellcolor[HTML]{EFEFEF}Vehicles} &
  \multicolumn{2}{c}{\cellcolor[HTML]{EFEFEF}COVID-19} &
  \multicolumn{2}{c}{\cellcolor[HTML]{EFEFEF}Vehicles} &
  \multicolumn{2}{c}{\cellcolor[HTML]{EFEFEF}COVID-19} &
  \multicolumn{2}{c}{\cellcolor[HTML]{EFEFEF}Vehicles} \\ \hline
\rowcolor[HTML]{EFEFEF} 
\cellcolor[HTML]{EFEFEF}\textbf{Seed Topics} &
  \multicolumn{2}{c}{\cellcolor[HTML]{EFEFEF}\begin{tabular}[c]{@{}c@{}}Side effects  \\ Ineffective\end{tabular}} &
  \multicolumn{2}{c}{\cellcolor[HTML]{EFEFEF}\begin{tabular}[c]{@{}c@{}}Autos\\ Motorcycles\end{tabular}} &
  \multicolumn{2}{c}{\cellcolor[HTML]{EFEFEF}\begin{tabular}[c]{@{}c@{}}Side effects  \\ Ineffective\end{tabular}} &
  \multicolumn{2}{c}{\cellcolor[HTML]{EFEFEF}\begin{tabular}[c]{@{}c@{}}Autos\\ Motorcycles\end{tabular}} &
  \multicolumn{2}{c}{\cellcolor[HTML]{EFEFEF}\begin{tabular}[c]{@{}c@{}}Side effects  \\ Ineffective\end{tabular}} &
  \multicolumn{2}{c}{\cellcolor[HTML]{EFEFEF}\begin{tabular}[c]{@{}c@{}}Autos\\ Motorcycles\end{tabular}} \\ \hline
\rowcolor[HTML]{FFFFFF} 
\textbf{Adherence\%} &
  50\% &
  100\% &
  1\% &
  99\% &
  35\% &
  {\color[HTML]{333333} 100\%} &
  0\% &
  94\% &
  {\color[HTML]{333333} 10\%} &
  {\color[HTML]{333333} 97\%} &
  \cellcolor[HTML]{FFFFFF}{\color[HTML]{333333} 1\%} &
  {\color[HTML]{333333} 94\%} \\
\rowcolor[HTML]{FFFFFF} 
\textbf{Hallucination \%} &
  5\% &
  0\% &
  9\% &
  0\% &
  0\% &
  0\% &
  2\% &
  0\% &
  {\color[HTML]{333333} 14\%} &
  {\color[HTML]{333333} 3\%} &
  \cellcolor[HTML]{FFFFFF}{\color[HTML]{333333} 10\%} &
  {\color[HTML]{333333} 3\%} \\
\rowcolor[HTML]{FFFFFF} 
\textbf{Aligned\%} &
  45\% &
  0\% &
  90\% &
  1\% &
  65\% &
  0\% &
  98\% &
  6\% &
  {\color[HTML]{333333} 76\%} &
  {\color[HTML]{333333} 0\%} &
  {\color[HTML]{333333} 89\%} &
  {\color[HTML]{333333} 3\%} \\ \hline
\rowcolor[HTML]{EFEFEF} 
\cellcolor[HTML]{EFEFEF}\textbf{Seed Topics} &
  \multicolumn{2}{c}{\cellcolor[HTML]{EFEFEF}No Seeds} &
  \multicolumn{2}{c}{\cellcolor[HTML]{EFEFEF}No Seeds} &
  \multicolumn{2}{c}{\cellcolor[HTML]{EFEFEF}No Seeds} &
  \multicolumn{2}{c}{\cellcolor[HTML]{EFEFEF}No Seeds} &
  \multicolumn{2}{c}{\cellcolor[HTML]{EFEFEF}No Seeds} &
  \multicolumn{2}{c}{\cellcolor[HTML]{EFEFEF}No Seeds} \\ \hline
\rowcolor[HTML]{FFFFFF} 
\cellcolor[HTML]{FFFFFF}\textbf{Adherence\%} &
  29\% &
  97\% &
  0\% &
  60\% &
  28\% &
  99\% &
  0\% &
  50\% &
  7\% &
  85\% &
  0\% &
  59\% \\
\rowcolor[HTML]{FFFFFF} 
\cellcolor[HTML]{FFFFFF}\textbf{Hallucination \%} &
  13\% &
  3\% &
  2\% &
  8\% &
  36\% &
  1\% &
  2\% &
  7\% &
  46\% &
  15\% &
  5\% &
  26\% \\
\rowcolor[HTML]{FFFFFF} 
\cellcolor[HTML]{FFFFFF}\textbf{Aligned\%} &
  58\% &
  0\% &
  98\% &
  32\% &
  36\% &
  0\% &
  98\% &
  43\% &
  47\% &
  0\% &
  95\% &
  15\% \\ \hline
\end{tabular}
}
\caption{ Direct comparison between TopicMistral and off-the-shelf Mistral across different combinations of document domains, granularity descriptions, and seed topics. Note that the testing samples used are strictly excluded from the DPO training samples. TM-7B denotes TopicMistral.}
\label{tab:hallucinations_2_dpo}
\end{table*}

\begin{table*}[!t]
\resizebox{\textwidth}{!}{%
\begin{tabular}{lclclclclclcl}
\hline
\rowcolor[HTML]{FFCE93} 
\textbf{Models} &
  \textbf{M-7B} &
  \multicolumn{1}{c}{\cellcolor[HTML]{FFCE93}\textbf{TM-7B}} &
  \textbf{M-7B} &
  \multicolumn{1}{c}{\cellcolor[HTML]{FFCE93}\textbf{TM-7B}} &
  \textbf{M-7B} &
  \multicolumn{1}{c}{\cellcolor[HTML]{FFCE93}\textbf{TM-7B}} &
  \textbf{M-7B} &
  \multicolumn{1}{c}{\cellcolor[HTML]{FFCE93}\textbf{TM-7B}} &
  \textbf{M-7B} &
  \multicolumn{1}{c}{\cellcolor[HTML]{FFCE93}\textbf{TM-7B}} &
  \textbf{M-7B} &
  \multicolumn{1}{c}{\cellcolor[HTML]{FFCE93}\textbf{TM-7B}} \\ \hline
\rowcolor[HTML]{EFEFEF} 
\textbf{Docs. / Gran. Desc.} &
  \multicolumn{4}{c}{\cellcolor[HTML]{EFEFEF}Sports News / Sports} &
  \multicolumn{4}{c}{\cellcolor[HTML]{EFEFEF}Computer News / Computer} &
  \multicolumn{4}{c}{\cellcolor[HTML]{EFEFEF}Politics News / Politics} \\ \cline{2-13} 
\rowcolor[HTML]{EFEFEF} 
\cellcolor[HTML]{EFEFEF}\textbf{Seed Topics} &
  \multicolumn{2}{c}{\cellcolor[HTML]{EFEFEF}\begin{tabular}[c]{@{}c@{}}Baseball\\ Hocky\end{tabular}} &
  \multicolumn{2}{c}{\cellcolor[HTML]{EFEFEF}No Seeds} &
  \multicolumn{2}{c}{\cellcolor[HTML]{EFEFEF}\begin{tabular}[c]{@{}c@{}}Hardware \\ Software\end{tabular}} &
  \multicolumn{2}{c}{\cellcolor[HTML]{EFEFEF}No Seeds} &
  \multicolumn{2}{c}{\cellcolor[HTML]{EFEFEF}\begin{tabular}[c]{@{}c@{}}Government \\ Politician\end{tabular}} &
  \multicolumn{2}{c}{\cellcolor[HTML]{EFEFEF}No Seeds} \\ \hline
\rowcolor[HTML]{FFFFFF} 
\textbf{True Positive \%} &
  \multicolumn{1}{l}{\cellcolor[HTML]{FFFFFF}68\%} &
  70\% &
  \multicolumn{1}{l}{\cellcolor[HTML]{FFFFFF}100\%} &
  81\% &
  \multicolumn{1}{l}{\cellcolor[HTML]{FFFFFF}100\%} &
  {\color[HTML]{333333} 93\%} &
  \multicolumn{1}{l}{\cellcolor[HTML]{FFFFFF}100\%} &
  \cellcolor[HTML]{FFFFFF}100\% &
  \multicolumn{1}{l}{\cellcolor[HTML]{FFFFFF}{\color[HTML]{333333} 28\%}} &
  {\color[HTML]{333333} 65\%} &
  \multicolumn{1}{l}{\cellcolor[HTML]{FFFFFF}{\color[HTML]{333333} 100\%}} &
  {\color[HTML]{333333} 70\%} \\ \hline
\end{tabular}
}
\caption{ Comparative experimental results on True Positive\% across three test sets. TM-7B denotes TopicMistral.}
\label{tab:hallucinations_3_true_positive}
\end{table*}

\label{hallu_results}

This section investigates the hallucination effect in LLM-based topic modelling. We first introduce the hallucination evaluation metrics (Section~\ref{sec:hallu_metric}), followed by results and discussion in Section~\ref{sec:hallu_results}.

\subsection{Hallucinated Topics Evaluation Metrics} \label{sec:hallu_metric}
To investigate the degree of LLM hallucination, we manually conduct evaluations on the different prompt settings as shown in Table \ref{tab:hallucinations} and Table \ref{tab:hallucinations_2_dpo}. \footnote{The manual evaluations were conducted by the authors of this paper.} Since each of our hallucination test sets contains 100 samples, we report three metrics expressed as percentages.
\paragraph{Instruction Adherence Rate (Adherence\%)}: This metric demonstrates that the topics generated are aligned with human instructions. For example, given a sports news article, the LLM should return answers like `No related topics' if we instruct it to generate COVID-19 related topics.

\paragraph{Hallucination Rate\%} Measures the cases where topics unrelated to the given text are hallucinated topics. An example is when we instruct LLMs to generate topics related to COVID-19 given a sports news article; however, the LLMs return topics related to COVID-19 instead of `No related topics'.

\paragraph{Task-Aligned Rate (Aligned\%)} Similarly, Aligned\% shows the number of cases where the model generates topics related to the given text but does not align with the instruction provided.

\paragraph{True Positive\%} displays the ratio at which LLMs correctly return topics aligned with instructions and the given document. We use this metric to evaluate whether our fine-tuned model can accurately determine if there are no related topics in the given text, rather than defaulting to the answer `No related topics'. For instance, we assess our new model by providing sports-related news articles and guiding it to generate topics associated with sports.

\subsection{Results and Discussion} \label{sec:hallu_results}
\paragraph{OpenAI's GPT generates significantly fewer hallucinated topics than those open source LLMs} We uncover that GPT closely adheres to our instructions regarding the domain of the desired topics, with a mean adherence rate of 99.3\%, whereas most vanilla open-source LLMs tend to overlook human instructions (achieving a mean adherence rate of only 8.6\% across all settings). For example, GPT can usually return `No related topics' if there are no relevant topics in the given text. This suggests that most open source LLMs are at risk of generating hallucinated topics, especially when provided with adversarial prompts. Meanwhile, this finding highlights the importance of our research.\footnote{Top 10 frequent topics across all LLMs and prompt strategies are provided in the appendix.}

\paragraph{Seed topics play an important role in guiding LLMs to generate the expected topics.} For all off-the-shelf LLMs (see Table \ref{tab:hallucinations}), we observe that using an adversarial Baseline prompt (w/o Seed Topics) can result in a higher number of hallucinated topics (average hallucination rates of 8\% vs. 18\% across 4 off-the-shelf LLMs). Even GPT can occasionally generate hallucinated topics when given an adversarial Baseline prompt without seed topics (1\%).

\paragraph{In most cases, unwanted topics are related to the given text rather than the granularity of the description.} LLMs tend to neglect the human-designed granularity descriptions and seed topics, instead focusing on the task of topic modelling based on the given document. (see Aligned\% scores in Table \ref{tab:hallucinations_2_dpo}

\paragraph{TopicMistral demonstrates significantly better capability in generating fewer hallucinated topics when provided with an adversarial prompt.} As shown in Table \ref{tab:hallucinations_2_dpo}, TopicMistral achieves a low average hallucination rate of $1\%$ versus $7\%$ and a higher adherence rate of $97\%$ versus 16\%, compared to off-the-shelf Mistral. However, when given a granularity description prompt without seed topics, our TopicMistral can sometimes produce hallucinated topics, with an average rate of $10\%$. This further reinforces the finding that using seed topics is an effective hallucination mitigation approach.

\paragraph{TopicMistral does not default to producing `No related topics' when provided with a non-adversarial prompt.} In Table \ref{tab:hallucinations_3_true_positive}, we display the results of comparative experiments by using related granularity descriptions and seed topics. We find that the fine-tuned TopicMistral is sensitive to documents lacking content related to the granularity description and seed topics, which results in the generation of `No related topics'. However, it still achieves better performance than the off-the-shelf Mistral (the average True Positive \% of $76\%$ v.s. $68\%$).
Note that off-the-shelf Mistral always generates topics that exist in the given text when no seed topic is provided; therefore, we assume that the True Positive value will be 100\%.

\begin{figure}[!t]
    \centering
    \includegraphics[width=1.1\columnwidth]{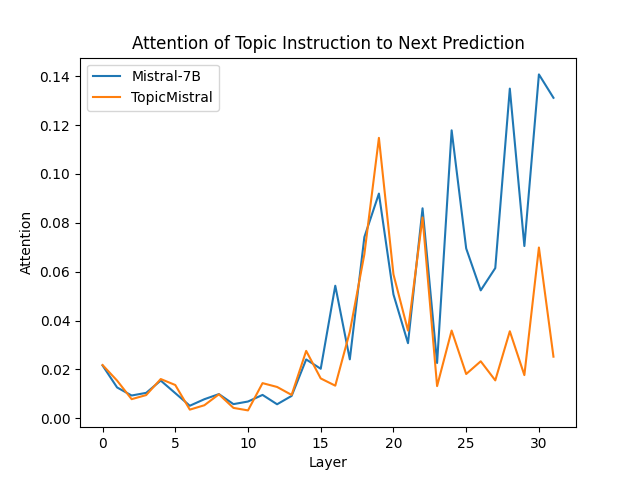} 
    \caption{Average attention weight shift on Mistral-7B and TopicMistral. The x-axis represents the number of hidden layers and y-axis indicates the average attention weight from topic instruction to the next prediction token.}
    \label{fig:attention} 
\end{figure}

\paragraph{TopicMistral aggregates less hallucinated topic information during the attention propagation within the deep layers.} To further investigate the freasons behind TopicMistral generating fewer hallucinations, we conduct an analysis of the attention weight shift \citep{wang2023label} from the topic instruction (i.e., seed topic tokens) to the next prediction token within the hidden layers of TopicMistral and off-the-shelf Mistral. Figure \ref{fig:attention} shows the average attention weight shift in instances of topic hallucinations using Mistral-7B. We observe that the attention weights on TopicMistral significantly decrease in the deep layers, indicating a reduced aggregation of hallucinated topic information for the next prediction token.

\section{Conclusion}
This work focused on improving LLM performance on topic modelling. 
The novel contributions of the paper lie in introducing novel approaches to address topic granularity and hallucination issues that occur when off-the-shelf API-based and open-source LLMs are used for LLM-driven topic modelling. 
We also demonstrate an efficient LLM fine-tuning pipeline through the implementation of the DPO fine-tuning framework. 
Our method, which does not require human annotation and favours an automated pipeline, has proven effective in fine-tuning LLM outputs to produce a higher number of coherent and relevant topics while simultaneously reducing the number of hallucinated topics. 

\section*{Ethics Statement}
Our work has been approved by the Research Ethics Committee of our institute. All datasets are publicly available via the links or the respective cited papers.


\bibliography{custom}
\appendix
\section*{Appendix}

\begin{table*}[]
\resizebox{\textwidth}{!}{%
\begin{tabular}{
>{\columncolor[HTML]{EFEFEF}}l |l}
\hline
\cellcolor[HTML]{FFCE93}\textbf{\#} &
  \multicolumn{1}{c}{\cellcolor[HTML]{FFCE93}\textbf{20NG Top 10 Topics}} \\ \hline
\textbf{\begin{tabular}[c]{@{}l@{}}Original \\ Categories\end{tabular}} &
  \cellcolor[HTML]{EFEFEF}\begin{tabular}[c]{@{}l@{}}Computer {[}graphics, os.ms-windows.misc, sys.ibm.pc.hardware, sys.mac.hardware, windows.x, misc.forsale{]},\\ Recreation. {[}autos, motorcycles, sport.baseball, sport.hockey{]}, Science {[}electronics, medical, space, crypt{]},\\ Social {[}religion.christian{]}, Talk. {[}politics.guns, politics.mideast, politics.misc, religion.misc{]}, alt.atheism\end{tabular} \\ \hline
\textbf{TopicMistral} &
  \cellcolor[HTML]{EFEFEF}Computer, Sports, Religion, Politics, Technology, Medicine, Security, Car, Software, Space \\ \hline
\textbf{\begin{tabular}[c]{@{}l@{}}TopicMistral \\ Dynamic\end{tabular}} &
  \cellcolor[HTML]{EFEFEF}Technology, Computer, Religion, Sports, Politics, Medicine, Security, Car, Software, Space \\ \hline
\textbf{\begin{tabular}[c]{@{}l@{}}Mistral 7B \\ Baseline\end{tabular}} &
  Baseball, Homosexuality, Hockey, Email, Sale, Religion, Windows, Car, Christianity, Faith \\ \hline
\textbf{\begin{tabular}[c]{@{}l@{}}Mistral 7B \\ Gran. Desc.\end{tabular}} &
  Sports, Religion, Baseball, Politics, Hockey, Business, Homosexuality, Technology, Space, Software, Cars, Genocide \\ \hline
\textbf{\begin{tabular}[c]{@{}l@{}}Mistral 7B \\ Seed Topics\end{tabular}} &
  Baseball, Hardware, Religion, Hockey, News, Sports, Politics, Software, Genocide, Cars, Bible, Encryption \\ \hline
\textbf{\begin{tabular}[c]{@{}l@{}}LLaMA 7B\\ Baseline\end{tabular}} &
  Book, God, Encryption, Religion, Government, Technology, Faith, Genocide, Window, Do \\ \cline{2-2} 
\textbf{\begin{tabular}[c]{@{}l@{}}LLaMA 7B\\ Gran. Desc.\end{tabular}} &
  Religion, Baseball, Political scandal, Technology, Economic downturn, Hockey, News article, Sport, Encryption, Sale \\ \cline{2-2} 
\textbf{\begin{tabular}[c]{@{}l@{}}LLaMA 7B\\ Seed Topics\end{tabular}} &
  Hardware, Technology, Sport, Religion, Baseball, Car, Software, Hockey, Windows, Politics \\ \hline
\textbf{\begin{tabular}[c]{@{}l@{}}LLaMA 13B \\ Baseline\end{tabular}} &
  Baseball, Homosexuality, Space exploration, Hockey, Fire, Religion, Morality, Genocide, Christianity, Encryption \\ \hline
\textbf{\begin{tabular}[c]{@{}l@{}}LLaMA 13B \\ Gran. Desc.\end{tabular}} &
  Technology, Computer hardware, Hockey, Religion, Baseball, Space exploration, Hardware, Sports, Sales, Political scandals \\ \hline
\textbf{\begin{tabular}[c]{@{}l@{}}LLaMA 13B\\ Seed Topics\end{tabular}} &
  Technology, Baseball, Hockey, Hardware, Sports, Space exploration, Religion, Computer hardware, Motorcycles, Sales \\ \hline
\textbf{\begin{tabular}[c]{@{}l@{}}GPT\\ Baseline\end{tabular}} &
  Encryption, Homosexuality, Christianity, Expertise, Expertise in, Price, Genocide, Original sin, Motorcycle, Modem \\ \hline
\textbf{\begin{tabular}[c]{@{}l@{}}GPT\\ Gran. Desc.\end{tabular}} &
  Sport, Politics, Entertainment, Encryption, Technology, Current event, Law enforcement, Homosexuality, Morality, Hockey \\ \hline
\textbf{\begin{tabular}[c]{@{}l@{}}GPT\\ Seed Topics\end{tabular}} &
  \begin{tabular}[c]{@{}l@{}}Technology, Politics, Entertainment, Encryption, Computer hardware, Environment, Baseball players, Baseball, Hockey, \\ Armenian genocide\end{tabular} \\ \hline
\cellcolor[HTML]{FFCE93}\textbf{\#} &
  \multicolumn{1}{c}{\cellcolor[HTML]{FFCE93}\textbf{20NG Specific Domain Top 10 Topics}} \\ \hline
\textbf{\begin{tabular}[c]{@{}l@{}}Original \\ Categories\end{tabular}} &
  \cellcolor[HTML]{EFEFEF}Recreation. {[}autos, motorcycles, sport.baseball, sport.hockey{]} \\ \hline
\cellcolor[HTML]{EFEFEF}\textbf{TopicMistral} &
  \cellcolor[HTML]{EFEFEF}Baseball, Hockey, Car, Sports, Cars, Football, Motor, Bike, Insurance, Engine \\ \hline
\cellcolor[HTML]{EFEFEF}\textbf{\begin{tabular}[c]{@{}l@{}}TopicMistral \\ Dynamic\end{tabular}} &
  \cellcolor[HTML]{EFEFEF}Sports, Baseball, Recreation, Cycling, Ice, Hockey, Football, Car, Bicycle \\ \hline
\cellcolor[HTML]{EFEFEF}\textbf{\begin{tabular}[c]{@{}l@{}}Mistral 7B \\ Baseline\end{tabular}} &
  Baseball, Hockey, Nhl, Game, Espn, Bike, Engine, Teams, Car \\ \hline
\cellcolor[HTML]{EFEFEF}\textbf{\begin{tabular}[c]{@{}l@{}}Mistral 7B \\ Gran. Desc.\end{tabular}} &
  Baseball, Sports, Hockey, Motorcycles, Nhl, Politics, Ice hockey, Cars, Car, Sports news \\ \hline
\cellcolor[HTML]{EFEFEF}\textbf{\begin{tabular}[c]{@{}l@{}}Mistral 7B \\ Seed Topics\end{tabular}} &
  Baseball, Hockey, Sports, Nhl, Politics, Motorcycles, Ice hockey, Bike, Espn, Football \\ \hline
\textbf{\begin{tabular}[c]{@{}l@{}}LLaMA 7B\\ Baseline\end{tabular}} &
  Hockey, Baseball, Bruins, Espn, Bike, Bmw, Sabres, Playoffs, Players, Time \\ \hline
\textbf{\begin{tabular}[c]{@{}l@{}}LLaMA 7B\\ Gran. Desc.\end{tabular}} &
  Hockey, Baseball, Sports news, Motorcycles, Injury, Racing, Playoffs, Car maintenance, Cars, Sabres \\ \hline
\textbf{\begin{tabular}[c]{@{}l@{}}LLaMA 7B\\ Seed Topics\end{tabular}} &
  Hockey, Baseball, Sports, Cars, Motorcycles, Safety, Pitching, Playoffs, Car maintenance \\ \hline
\cellcolor[HTML]{EFEFEF}\textbf{\begin{tabular}[c]{@{}l@{}}LLaMA 13B \\ Baseline\end{tabular}} &
  Baseball, Hockey, Bike, Motorcycles, Injury, Espn, Waving, Price, Bruins, Space exploration \\ \hline
\cellcolor[HTML]{EFEFEF}\textbf{\begin{tabular}[c]{@{}l@{}}LLaMA 13B \\ Gran. Desc.\end{tabular}} &
  \begin{tabular}[c]{@{}l@{}}Baseball, Hockey, Sports, Motorcycles, Player performance, Car maintenance, Celebrity gossip, Political scandals, \\ Cars, Automotive\end{tabular} \\ \hline
\cellcolor[HTML]{EFEFEF}\textbf{\begin{tabular}[c]{@{}l@{}}LLaMA 13B\\ Seed Topics\end{tabular}} &
  \begin{tabular}[c]{@{}l@{}}Baseball, Hockey, Sports, Motorcycles, Car maintenance, Cars, Player performance, Pitching, Hockey playoffs, \\ Motorcycle safety\end{tabular} \\ \hline
\cellcolor[HTML]{EFEFEF}\textbf{\begin{tabular}[c]{@{}l@{}}GPT\\ Baseline\end{tabular}} &
  Baseball, Hockey, Grant fuhr, Expertise, Baseball game, Countersteering, Bike, Hockey coverage, Expertise in, Motorcycles \\ \hline
\cellcolor[HTML]{EFEFEF}\textbf{\begin{tabular}[c]{@{}l@{}}GPT\\ Gran. Desc.\end{tabular}} &
  Sports news, Sports events, Baseball, Hockey, Basketball, Football, Tennis, Sports, Sports updates, News \\ \hline
\cellcolor[HTML]{EFEFEF}\textbf{\begin{tabular}[c]{@{}l@{}}GPT\\ Seed Topics\end{tabular}} &
  Baseball, Hockey, Entertainment, Sports, Politics, Technology, Car maintenance, Grant fuhr, Pittsburgh penguins, Playoffs \\ \hline
\end{tabular}
}
\caption{Top 10 most frequent topics across all prompt strategies and LLMs on the 20NG and 20NG Specific datasets.}
\label{tab:top_10 20ngs}
\end{table*}

\begin{table*}[]
\resizebox{\textwidth}{!}{%
\begin{tabular}{
>{\columncolor[HTML]{EFEFEF}}l |l}
\hline
\cellcolor[HTML]{FFCE93}\textbf{\#} &
  \multicolumn{1}{c}{\cellcolor[HTML]{FFCE93}\textbf{Wiki Top 10 Topics}} \\ \hline
\textbf{\begin{tabular}[c]{@{}l@{}}Original \\ Categories\end{tabular}} &
  \cellcolor[HTML]{EFEFEF}\begin{tabular}[c]{@{}l@{}}Media and drama, Agriculture, food, and drink, Philosophy and religion, Warfare, Engineering and technology, \\ Art and architecture, Mathematics, Geography and places, History, Social sciences and society, Video games, \\ Music, Language and literature, Sports and recreation, Natural sciences\end{tabular} \\ \hline
\textbf{TopicMistral} &
  \cellcolor[HTML]{EFEFEF}History, Music, Sports, Technology, Television, Politics, Military, Football, Transport, Geography \\ \hline
\textbf{\begin{tabular}[c]{@{}l@{}}TopicMistral \\ Dynamic\end{tabular}} &
  \cellcolor[HTML]{EFEFEF}History, Music, Technology, Television, Sports, Military, Politics, Football, Transport, Film \\ \hline
\textbf{\begin{tabular}[c]{@{}l@{}}Mistral 7B \\ Baseline\end{tabular}} &
  History, Music, Football, Politics, Film, Early life, Meteorological history, World War II, Gameplay, Architecture \\ \hline
\textbf{\begin{tabular}[c]{@{}l@{}}Mistral 7B \\ Seed Topics\end{tabular}} &
  History, Music, Football, Politics, Meteorological history, Early life, Architecture, , Film, World War II \\ \hline
\textbf{\begin{tabular}[c]{@{}l@{}}LLaMA 7B\\ Baseline\end{tabular}} &
  History, Early life, Route description, Gameplay, Design, Music, Meteorological history, World War II, Geography \\ \hline
\textbf{\begin{tabular}[c]{@{}l@{}}LLaMA 7B\\ Seed Topics\end{tabular}} &
  History, Music, Football, Architecture, Politics, World War II, Early life, Boat Race, Geography \\ \hline
\textbf{\begin{tabular}[c]{@{}l@{}}LLaMA 13B \\ Baseline\end{tabular}} &
  History, Music, Geography, Football, Architecture, Transportation, World War II, World War I, Sports \\ \hline
\textbf{\begin{tabular}[c]{@{}l@{}}LLaMA 13B\\ Seed Topics\end{tabular}} &
  History, Music, Transportation, Sports, Geography, Military, Football, Architecture, Education \\ \hline
\textbf{\begin{tabular}[c]{@{}l@{}}GPT\\ Baseline\end{tabular}} &
  \begin{tabular}[c]{@{}l@{}}History, Music, World War II, Meteorological history, Early life, Geography, Football career, Music Production, \\ Personal life, Military Service\end{tabular} \\ \hline
\textbf{\begin{tabular}[c]{@{}l@{}}GPT\\ Seed Topics\end{tabular}} &
  \begin{tabular}[c]{@{}l@{}}History, Music, Transportation, Naval History, World War II, Geography, Meteorological history, Education, \\ Performance, Relationships\end{tabular} \\ \hline
\cellcolor[HTML]{FFCE93}\textbf{\#} &
  \multicolumn{1}{c}{\cellcolor[HTML]{FFCE93}\textbf{Bills Top 10 Topics}} \\ \hline
\textbf{\begin{tabular}[c]{@{}l@{}}Original \\ Categories\end{tabular}} &
  \cellcolor[HTML]{EFEFEF}\begin{tabular}[c]{@{}l@{}}Defense, Agriculture, Public Lands, Culture, Macroeconomics, Technology, Law and Crime, Health, Domestic \\ Commerce, Transportation, Energy, Social Welfare, Immigration, Foreign Trade, Civil Rights, Government \\ Operations, Education, Housing, Environment, International Affairs, Labor\end{tabular} \\ \hline
\cellcolor[HTML]{EFEFEF}\textbf{TopicMistral} &
  \cellcolor[HTML]{EFEFEF}Government, Security, Education, Finance, Regulation, Health, Tax, Economy, Energy, Environment \\ \hline
\cellcolor[HTML]{EFEFEF}\textbf{\begin{tabular}[c]{@{}l@{}}TopicMistral \\ Dynamic\end{tabular}} &
  \cellcolor[HTML]{EFEFEF}Government, Security, Regulation, Tax, Economy, Medicare, Finance, Education, Energy, Medicine \\ \hline
\cellcolor[HTML]{EFEFEF}\textbf{\begin{tabular}[c]{@{}l@{}}Mistral 7B \\ Baseline\end{tabular}} &
  \begin{tabular}[c]{@{}l@{}}Tariff Schedule, Act, Tax Credit, Medicare, Duty Suspension, Education, Immigration, Veterans, Social Security, \\ Amendment\end{tabular} \\ \hline
\cellcolor[HTML]{EFEFEF}\textbf{\begin{tabular}[c]{@{}l@{}}Mistral 7B \\ Seed Topics\end{tabular}} &
  Education, Tariffs, Medicare, Tax Credit, Health, Immigration, Agriculture, Social Security, Veterans, Funding \\ \hline
\textbf{\begin{tabular}[c]{@{}l@{}}LLaMA 7B\\ Baseline\end{tabular}} &
  Medicare, Tariff, Tax Credit, Tariff Schedule, Education, Suspension, Duty Suspension, Immigration, Veterans \\ \hline
\textbf{\begin{tabular}[c]{@{}l@{}}LLaMA 7B\\ Seed Topics\end{tabular}} &
  Education, Healthcare, Trade, Immigration, Taxation, Medicare, Health, Tariff, Tax Credit \\ \hline
\cellcolor[HTML]{EFEFEF}\textbf{\begin{tabular}[c]{@{}l@{}}LLaMA 13B \\ Baseline\end{tabular}} &
  \begin{tabular}[c]{@{}l@{}}Tariffs, Medicare, Education, Tax Credits, Suspension, Tax Credit, Renewable Energy, Land Management, \\ Social Security\end{tabular} \\ \hline
\cellcolor[HTML]{EFEFEF}\textbf{\begin{tabular}[c]{@{}l@{}}LLaMA 13B\\ Seed Topics\end{tabular}} &
  \begin{tabular}[c]{@{}l@{}}Tariffs, Education, Medicare, Tax Credits, Trade, Environmental Protection, Immigration, Transportation, \\ Land Management\end{tabular} \\ \hline
\cellcolor[HTML]{EFEFEF}\textbf{\begin{tabular}[c]{@{}l@{}}GPT\\ Baseline\end{tabular}} &
  \begin{tabular}[c]{@{}l@{}}Tariff Schedule, Duty Suspension, Tax Credit, Veterans Affairs, Education, Environmental Protection, Immigration, \\ Taxation, Homeland Security, Tax Credits\end{tabular} \\ \hline
\cellcolor[HTML]{EFEFEF}\textbf{\begin{tabular}[c]{@{}l@{}}GPT\\ Seed Topics\end{tabular}} &
  \begin{tabular}[c]{@{}l@{}}Trade, Education, Taxation, Tax Credits, Environmental Protection, Immigration, Land Management, National\\  Security, Healthcare, Financial Regulation\end{tabular} \\ \hline
\end{tabular}
}
\caption{Top 10 most frequent topics across all prompt strategies and LLMs on Bills and Wiki datasets.}
\label{tab:top_10_wiki_bills}
\end{table*}

\end{document}